\documentclass{article}

 \PassOptionsToPackage{numbers, compress}{natbib}
\usepackage[final]{neurips_2024}
\usepackage{color}
\usepackage[dvipsnames]{xcolor}
\usepackage{colortbl}
\usepackage{graphicx}
\usepackage{booktabs}
\usepackage{xspace}
\usepackage[utf8]{inputenc} 
\usepackage[T1]{fontenc}    
\usepackage[colorlinks,
            linkcolor=VioletRed,      
            anchorcolor=VioletRed, 
            citecolor=VioletRed,       
            ]{hyperref}
\usepackage{subfigure}
\usepackage{tcolorbox}
\usepackage{wrapfig}
\usepackage{url}            
\usepackage{booktabs}       
\usepackage{amsfonts}       
\usepackage{nicefrac}       
\usepackage{microtype}      
\usepackage{xcolor}         
\usepackage{bm}
\usepackage{graphicx}
\usepackage{amsmath}
\usepackage{listings}
\usepackage{cleveref}
\usepackage{longtable}
\usepackage{geometry}
\usepackage{algorithm}
\usepackage{amsthm}
\usepackage{tikz}
\usepackage{multirow}
\usepackage{graphicx}
\definecolor{Gray}{gray}{0.85}
\newcommand{\Gray}[0]{\rowcolor{gray!20}}

\definecolor{sclgreyblue}{rgb}{0.2,0.3,0.5}%

\newcommand{\abbr}[0]{R1-Reward\xspace}

\usepackage{soul}

\usepackage{xcolor}
\definecolor{CFE2F3}{HTML}{CFE2F3}

\title{R1-Reward: Training Multimodal Reward Model Through Stable Reinforcement Learning}
\vspace{-0.6cm}
\author{
\vspace{-0.4cm}
\\ 
    Yi-Fan Zhang$^{1}$, Xingyu Lu$^{2}$, Xiao Hu$^{2}$, Chaoyou Fu$^{4,\dagger}$, Bin Wen$^{3,\spadesuit}$,Tianke Zhang$^{3}$\\ 
    Changyi Liu$^{3}$, Kaiyu Jiang$^{3}$, Kaibing Chen$^{3}$, Kaiyu Tang$^{3}$, Haojie Ding$^{3}$, Jiankang Chen$^{3}$\\
    Fan Yang$^{3}$, Zhang Zhang$^{1,\dagger}$, Tingting Gao$^{3}$, Di Zhang$^{3}$, Guorui Zhou$^{3}$, Liang Wang$^{1}$\\
    \\ 
    $^{1}$CASIA,  $^{2}$THU, $^{3}$KuaiShou, $^{4}$NJU
    \and
    \footnotesize{
    $^{\spadesuit}$~Project Leader \;
    $^{\dagger}$~Corresponding Author \;}
    \\ 
    \url{https://github.com/yfzhang114/r1_reward}
}

\begin{document}

\maketitle

\vspace{-0.6cm}
\begin{abstract}
Multimodal Reward Models (MRMs) play a crucial role in enhancing the performance of Multimodal Large Language Models (MLLMs). While recent advancements have primarily focused on improving the model structure and training data of MRMs, there has been limited exploration into the effectiveness of long-term reasoning capabilities for reward modeling and how to activate these capabilities in MRMs. In this paper, we explore how Reinforcement Learning (RL) can be used to improve reward modeling. Specifically, we reformulate the reward modeling problem as a rule-based RL task. However, we observe that directly applying existing RL algorithms, such as Reinforce++, to reward modeling often leads to training instability or even collapse due to the inherent limitations of these algorithms. To address this issue, we propose the StableReinforce algorithm, which refines the training loss, advantage estimation strategy, and reward design of existing RL methods. These refinements result in more stable training dynamics and superior performance. To facilitate MRM training, we collect 200K preference data from diverse datasets. Our reward model, R1-Reward, trained using the StableReinforce algorithm on this dataset, significantly improves performance on multimodal reward modeling benchmarks. Compared to previous SOTA models, R1-Reward achieves a $8.4\%$ improvement on the VL Reward-Bench and a $14.3\%$ improvement on the Multimodal Reward Bench. Moreover, with more inference compute, R1-Reward's performance is further enhanced, highlighting the potential of RL frameworks in optimizing MRMs.
\end{abstract}

\begin{figure}[h]
    \centering
    \includegraphics[width=\linewidth]{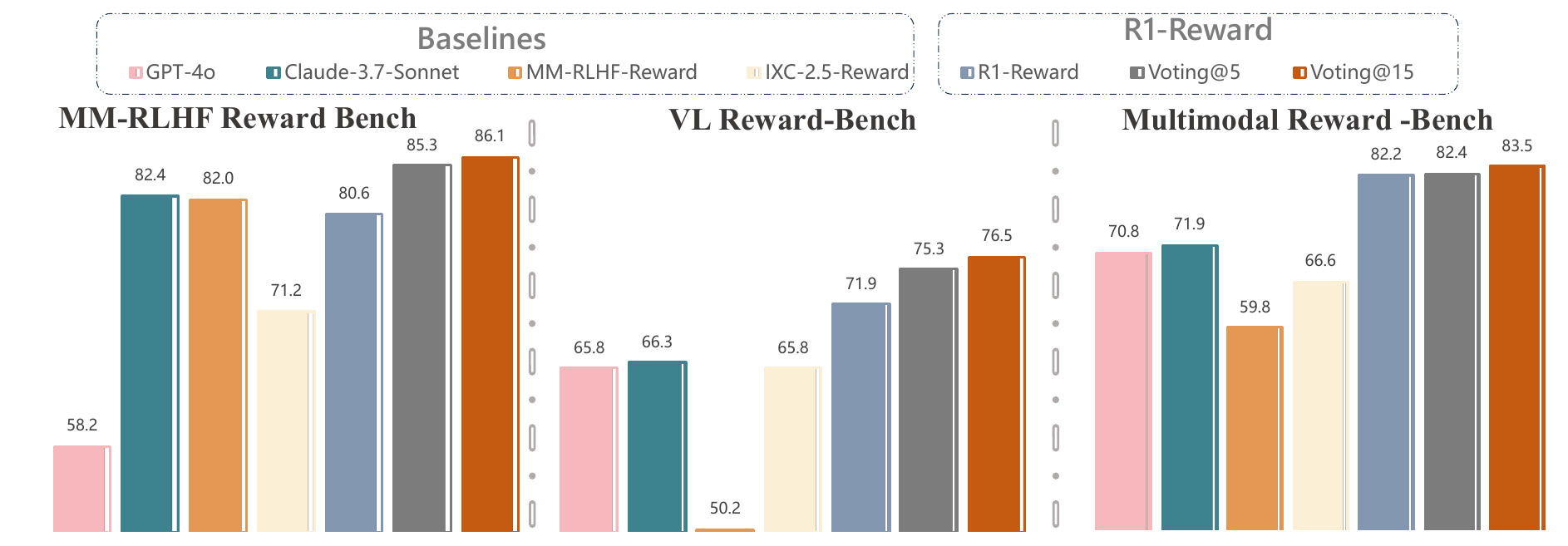}
    \vspace{-0.4cm}
    \caption{\textbf{R1-Reward performance on multimodal reward benchmarks.} Performance improves significantly when using a majority voting strategy (Voting@$5/15$) over multiple inference samples.}
     \vspace{-0.2cm}
    \label{fig:teaser}
\end{figure}
\vspace{-0.2cm}
\section{Introduction}
\begin{figure*}[t]
\subfigure[Policy Loss Convergence]{ 
\begin{minipage}[t]{0.49\linewidth}
\centering
 \includegraphics[width=\linewidth]{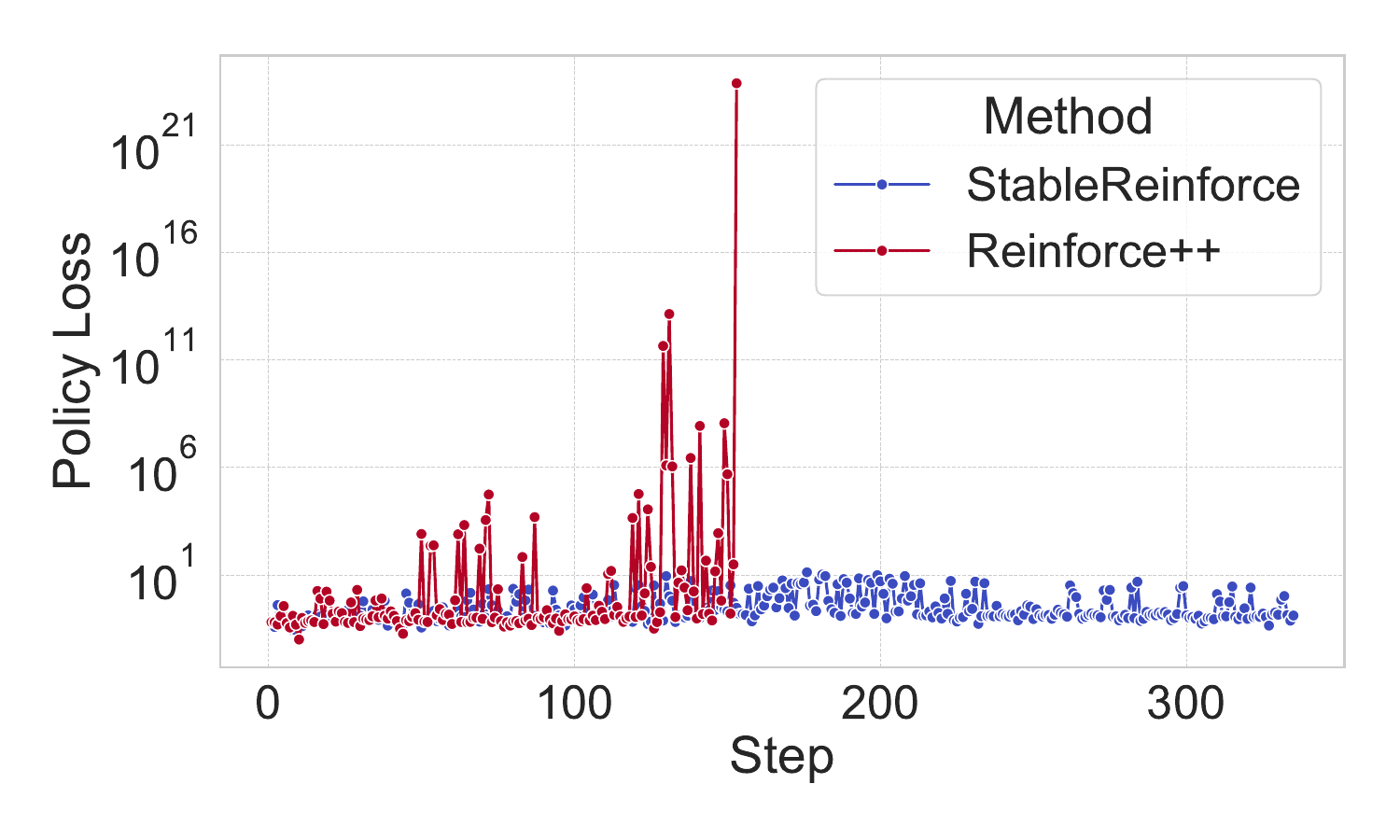}
\end{minipage}%
}%
\subfigure[Response Length During Training]{ 
\begin{minipage}[t]{0.49\linewidth}
\centering
 \includegraphics[width=\linewidth]{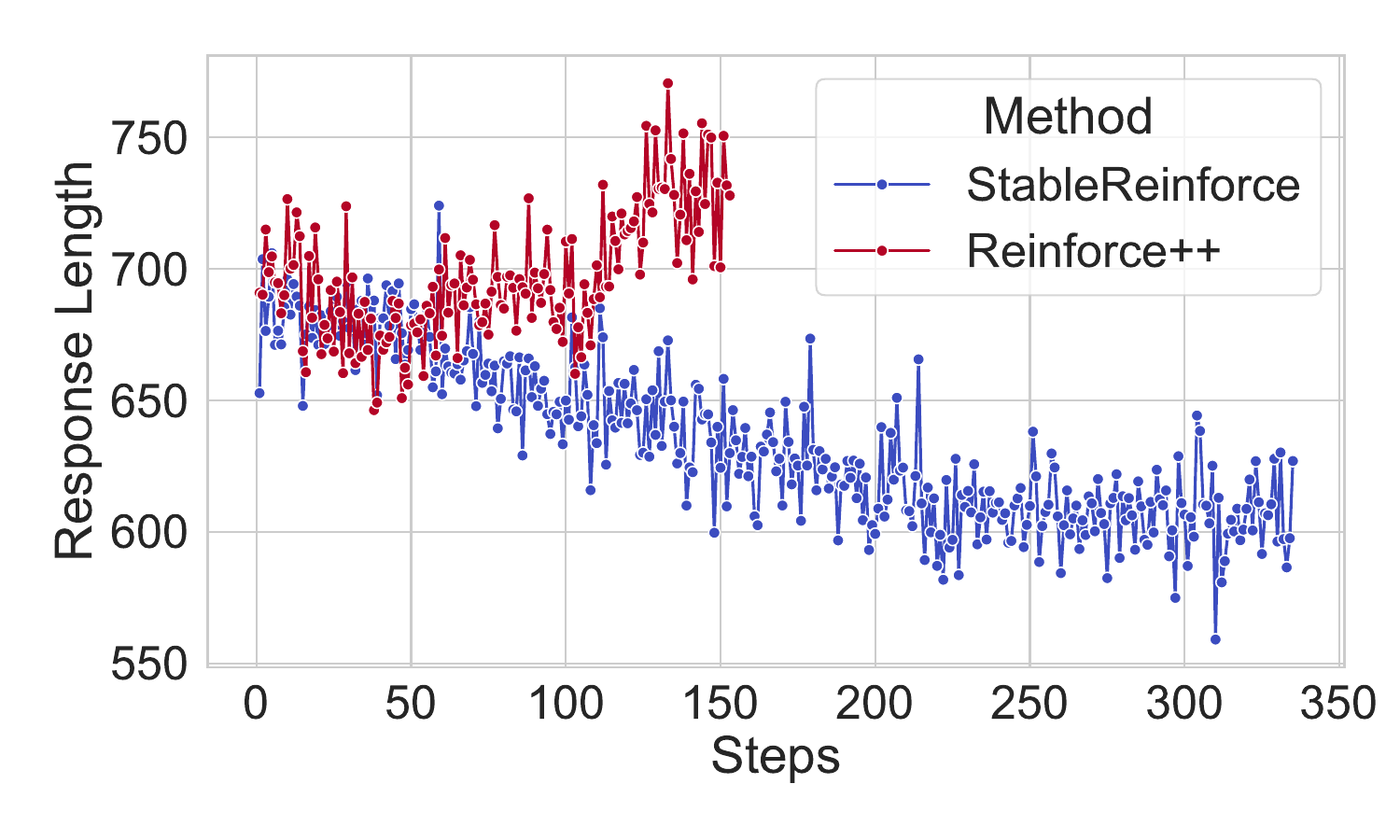}
\end{minipage}%
}%
\centering
\vspace{-0.2cm}
\caption{\textbf{Detailed comparison between StableReinforce and Reinforce++.} 
(a) StableReinforce exhibits faster and more stable convergence of the policy loss during training. 
(b)  StableReinforce continuously performs length compression, improving efficiency. 
Reinforce++ collapses around step 150, whereas StableReinforce remains stable, demonstrating its enhanced robustness. 
Additionally, after RL training with StableReinforce, the average response length is reduced by approximately 15\% compared to base model, suggesting potential improvements in reasoning token efficiency.}

\label{fig:res}
 \vspace{-0.4cm}
\end{figure*}

High-quality Multimodal Reward Models (MRMs)~\cite{pu2025judge,chen2024mllm,xiong2024llava,wang2025visualprmeffectiveprocessreward,zang2025internlm} play a crucial role in the development of Multimodal Large Language Models (MLLMs)~\cite{wang2024qwen2,deitke2024molmo,chen2024far,dai2024nvlm,agrawal2024pixtral,fu2025vita}. In the training phase, from an algorithmic perspective, the MRM provides reward signals for RL~\cite{2023llavarlhf,ouyang2022training}, directly influencing the stability and final outcomes of training. From a data perspective, a powerful MRM enables high-quality data filtering,  improving data quality by removing noisy samples ~\cite{zhang2025mm,lu2025dama}. In the inference phase, the MRM facilitates test-time scaling strategies, such as the best-of-N strategy, to select the optimal responses ~\cite{wang2025visualprmeffectiveprocessreward}. In the evaluation phase, a good MRM can serve as an evaluator to simplify the evaluation process, especially in open-ended scenarios~\cite{xiong2024llava}.

Recently, reinforcement learning~\cite{deepseekai2025deepseekr1incentivizingreasoningcapability,gpt-o1} has gained widespread application in the post-training process of MLLMs~\cite{yu2025aligning}, achieving remarkable improvements in traditional vision tasks~\cite{liu2025visual,shen2025vlm}, multimodal reasoning tasks~\cite{huang2025vision,peng2025lmm,meng2025mm}, video understanding tasks~\cite{feng2025video}, and omni-tasks~\cite{zhao2025r1}. Compared to traditional post-training strategies such as supervised fine-tuning and direct preference optimization~\cite{rafailov2023direct}, RL offers better generalization~\cite{chu2025sft} and demonstrates the ability to induce long-term reasoning capabilities~\cite{deepseekai2025deepseekr1incentivizingreasoningcapability}. However, recent improvements in MRMs have primarily focused on data~\cite{xiong2024llava,zang2025internlm} and structural aspects~\cite{zhang2025mm}, with little discussion on whether RL can be used to introduce long-term reasoning in order to improve multimodal reward modeling performance.

In this paper, we investigate whether RL algorithms can be applied to multimodal reward modeling tasks? Intuitively, the reward modeling problem can be transformed into a rule-based RL task, where the input consists of a given question and two answers. The target of the policy is to decide which answer is better. The reward during training can be obtained by comparing whether the model's judgment is consistent with the ground truth. Our goal is to enable the model to perform long-term reasoning and then provide the correct judgment. However, RL for reward modeling presents several unique challenges, and directly using traditional RL methods can easily cause training to collapse:

1. \textbf{Limitation of PPO~\cite{schulman2017proximal} and Related Algorithms~\cite{shao2024deepseekmathpushinglimitsmathematical}.} PPO and related algorithms rely on clipping the loss function to ensure training stability. However, we observe that when the advantage is negative and the current policy differs significantly from the reference policy, simple clipping fails to prevent instability, which may cause the training process to diverge or even crash.

2. \textbf{Instability of Advantage Normalization.} We observe that in the later stages of training, where the majority of rewards in a single batch are either 1 or 0 with very low variance, the commonly used advantage normalization technique (subtracting the mean and dividing by the variance) in algorithms such as GRPO~\cite{shao2024deepseekmathpushinglimitsmathematical} and Reinforce++~\cite{hu2025reinforce} can lead to extremely large or small advantage values for some samples. This can cause significant instability during training.


3. \textbf{Inconsistency Between Reasoning and Results.} During training, we frequently observe inconsistencies between the model's reasoning process and its final output. The model may judge one answer as better during reasoning but ultimately output an opposite answer. This happens because rule-based RL only scores the result without supervising the reasoning process, leading the model to learn to generate correct answers without coherent reasoning.

To this end, at the algorithm level, we propose StableReinforce, which introduces several key modifications to traditional RL methods. Specifically, we refine the clipping operation to mitigate numerical instability caused by large updates and introduce a robust advantage normalization technique that limits the impact of outliers. On the reward function design front, StableReinforce introduces a novel mechanism: the use of an MLLM as a referee. This referee evaluates the consistency between the model's reasoning process and the final result, ensuring that the reasoning aligns with the output. This consistency reward promotes more accurate and logically coherent decision-making.

During the training phase, directly training the MLLM using reinforcement learning yields suboptimal results. Therefore, a progressive difficulty training strategy is adopted. Initially, 200K preference data is collected from publicly available datasets, and GPT-4o generates corresponding thinking processes, referred to as R1-Reward-200K, to serve as cold-start SFT data. Meanwhile, for each sample, the number of sampling attempts GPT-4o requires to infer a conclusion matching the ground truth is recorded, which is considered the difficulty level of that sample. In the reinforcement learning phase, samples where GPT-4o requires at least two sampling attempts to arrive at the correct answer, or fails to answer correctly even after three attempts, are selected as training data. These samples are then used to train the model with the enhanced StableReinforce algorithm. As shown in Figure~\ref{fig:res}, the reinforcement learning phase effectively performs token compression, and also resulting in a noticeable performance improvement in our experiments.

R1-Reward performs excellently on common multimodal reward modeling benchmarks. As shown in Figure~\ref{fig:teaser}, R1-Reward outperforms the state-of-the-art (SOTA) on all the three benchmarks. Furthermore, R1-Reward exhibits strong inference time scalability. By sampling only five times and selecting the most frequent answer as the correct one, the accuracy of reward modeling improves substantially. On the MM-RLHF Reward Bench~\cite{zhang2025mm}, VL Reward-Bench~\cite{li2024vlrewardbenchchallengingbenchmarkvisionlanguage}, and Multimodal Reward Bench~\cite{yasunaga2025multimodal}, R1-Reward achieves improvements of $3.5\%$, $13.5\%$, and $14.6\%$, respectively, compared to SOTA. As the number of samples increases, performance continues to improve, demonstrating the potential of RL for multimodal reward modeling.

\vspace{-0.2cm}
\section{Related Work}
\vspace{-0.2cm}
\textbf{MLLMs.} Thanks to the success of language models, MLLMs have rapidly developed in recent years, with their task handling capabilities and model performance advancing at a fast pace~\cite{fu2024mme,zhang2024mme,yu2025aligning,li2022vision}. For example, traditional multi-modal large models perform well in handling complex high-resolution images and human dialogue~\cite{bai2025qwen25vltechnicalreport,gpt-o1,li2024llava,grattafiori2024llama3herdmodels,wang2024qwen2}. A series of works focus on improving the context length~\cite{shen2025longvitascalinglargemultimodal}, computational efficiency~\cite{zhang2024beyond,llavamini}, reducing hallucinations~\cite{lu2025dama,zhang2024debiasing}, enhancing conversational abilities~\cite{xiong2024llava}, and aligning with human preferences~\cite{zhang2025mm}. Omni-MLLMs are capable of simultaneously processing multiple modalities such as speech, video, images~\cite{li2025baichuan,zhao2025r1}, and even interacting with users via voice~\cite{fu2024vita,fu2025vita}. Unify-MLLMs can perform mixed-modal generation~\cite{xie2024show,team2024chameleon,xie2025mme}, for example, generating an image with auxiliary lines while understanding a math problem, enhancing both generation and comprehension abilities. Recently, with the success of Open AI's O1 model and Deepseek's R1 model~\cite{deepseekai2025deepseekr1incentivizingreasoningcapability}, the rule-based reinforcement learning approach has gained significant attention in the multi-modal field. Various studies are devoted to enhancing the reasoning capabilities of multi-modal models. However, as far as we know, no work has yet explored whether the reinforcement learning paradigm can be transferred into reward modeling.

\textbf{Reward Model Training.} The reward models most relevant to this paper are pure text reward models and multi-modal reward models. There are generally three main approaches to reward modeling. The first approach is to directly use a language model or multi-modal model as the reward model by designing precise prompts that allow them to output a score or ranking~\cite{xiong2024llava}. However, this method heavily depends on the model’s instruction-following ability and comprehension. The second approach involves connecting the latent representation of a language model to a reward head (typically an MLP or linear layer), where the model directly outputs a score. During training, the reward modeling is converted into a binary classification task. This approach is computationally efficient, but it lacks interpretability~\cite{liu2024skywork,zang2025internlm,INF-ORM-Llama3.1-70B,lou2024uncertainty,wang2024helpsteer2}. The final type of model simultaneously learns to evaluate the question-answer pair and creates an additional reward head to provide the score~\cite{yu2024self,zhang2025mm}. This model strikes a balance between interpretability and computational efficiency, but it usually requires specific data formats or training strategies. This paper proposes training a reward model through reinforcement learning. The model first outputs an inference for a given question-answer pair and ultimately provides a ranking. Through reinforcement learning, we force the model to learn the format of the reward modeling task, avoiding the shortcomings of the first approach without requiring an additional reward head, while maintaining the model’s interpretability.

\section{Preliminary and Limitations}
\subsection{Background and Limitations of Standard Reward Models}

Reward models are a key component for aligning model outputs with human preferences. Typically, a reward model starts with a pretrained LLM $\phi$, where the LLM head $h_l$ is replaced with a linear reward head $l_r$, enabling the model to output a scalar reward value. These models are trained using human-provided pairwise comparisons. Given a query $\mathbf{x}$, a preferred response $y_w$ and a less preferred response $y_l$, the reward model is optimized to assign higher rewards to preferred responses:
\begin{equation}
\ell_{\text{Reward}}(\theta) = 
\mathbb{E}_{\mathbf{x}, y_w, y_l} 
\Big[ 
    - \log \sigma \Big( r(y_w | \mathbf{x}) - r(y_l | \mathbf{x}) \Big)
\Big],
\end{equation}
where $r(y | \mathbf{x})$ is the scalar reward and $\sigma$ is the sigmoid function.

Despite their utility, standard reward models face significant limitations. First, they fail to fully utilize the rich and detailed feedback provided by high-quality human annotations, such as textual explanations and nuanced reasoning. Second, scalar rewards lack transparency, making it difficult for humans to understand how the reward is generated. These challenges highlight the need for a more interpretable and robust reward model that leverages critiques as intermediate reasoning steps.

\subsection{PPO and Reinforce++}
\textbf{Proximal Policy Optimization (PPO)}~\cite{schulman2017proximal} is a commonly used algorithm in RL that aims to optimize a policy directly while maintaining stable and efficient learning. PPO belongs to the family of policy gradient methods, where the objective is to improve the policy by maximizing the expected cumulative reward. Unlike traditional policy gradient methods, which can suffer from large updates and instability, PPO introduces a novel way to constrain policy updates, ensuring both efficient and stable learning. The objective function for PPO is defined as:

\[
L^{\text{PPO}}(\theta) = \frac{1}{|t|}\sum_t \left[ \min \left( \frac{\pi_\theta(a_t | s_t)}{\pi_{\theta_{\text{old}}}(a_t | s_t)} A_t, \, \text{clip} \left( \frac{\pi_\theta(a_t | s_t)}{\pi_{\theta_{\text{old}}}(a_t | s_t)}, 1 - \epsilon, 1 + \epsilon \right) A_t \right) \right]
\]

- \( \pi_\theta(a_t | s_t) \) is the probability of taking action \( a_t \) at state \( s_t \) under the current policy \( \theta \).

- \( \pi_{\theta_{\text{old}}}(a_t | s_t) \) is the probability under the old policy with parameters \( \theta_{\text{old}} \).

- \( A_t \) is the {advantage estimate} at time \( t \), which measures the relative desirability of the action taken.

- \( \epsilon \) is a small hyperparameter (typically \( 0.1 \leq \epsilon \leq 0.3 \)) that controls how much the policy can change.

The first term in the minimum represents the standard objective, while the second term applies a clipping mechanism. The clip function restricts the ratio of the new policy to the old policy to stay within the interval \( [1 - \epsilon, 1 + \epsilon] \). If the ratio exceeds this range, the objective is capped, preventing large updates that could destabilize the learning process.

PPO's key innovation is the introduction of a clipped objective function, which stabilizes the learning process by limiting the size of the policy updates. The method is both simple to implement and computationally efficient, making it a popular choice for a wide range of reinforcement learning tasks, including robotic control~\cite{singh2022reinforcement} and video game environments~\cite{shao2019survey}. 

\textbf{Reinforce++~\cite{hu2025reinforce} Enhancements.} Reinforce++ incorporates several key optimizations to enhance training stability and efficiency of PPO. One is the addition of a token-level Kullback-Leibler (KL) divergence penalty between the RL model and the supervised fine-tuning (SFT) model distributions. This token-level KL penalty is introduced into the reward function as follows:
$
r(s_t, a_t) = I(s_t = [EOS]) r(x, y) - \beta \text{KL}(t)
$
where \( x \) represents the input prompt, \( y \) denotes the generated response, \( I(s_t = [EOS]) \) is an indicator function that checks if the token \( t \) is the final token in the sequence (End of Sequence), and \( \beta \) is the KL penalty coefficient, controlling the strength of the regularization. 
Additionally, Reinforce++ introduces global batch-level reward normalization, clipping, and scaling for stability, as well as advantage normalization:
$
A_{\text{normalized}} = \frac{A - \mu_A}{\sigma_A}
$
Where \( \mu_A \) and \( \sigma_A \) are the mean and standard deviation of the advantage values.  REINFORCE++ is shown to be more stable compared to GRPO~\cite{shao2024deepseekmath} and faster than PPO~\cite{xie2025logic,cui2025process}.

\subsection{Drawbacks of Traditional PPO/Reinforce++}
During our training process, we observed two core issues in the Reinforce++ algorithms that can easily lead to model instability and poor performance, especially for reward model training.

\textbf{Instability Caused by Training Losses.} The typical PPO loss function is implemented as follows, given the log probabilities \(\log \pi_\theta(a_t | s_t)\), \(\log \pi_{\theta_{\text{old}}}(a_t | s_t)\), and advantages. The pseudocode in for calculating the loss is shown in Algorithm~\ref{alg:ppo_loss} (lines 0-3). If the ratio \(\frac{\pi_\theta(a_t | s_t)}{\pi_{\theta_{\text{old}}}(a_t | s_t)}\) differs significantly, two main issues arise. First, the expression \((\text{log\_probs} - \text{old\_log\_probs}).\exp()\) can lead to numerical instability. When the difference in token probabilities is large, the exponential function may overflow, causing the model to crash. Even if the computation proceeds normally, if the advantage is negative, \(-\text{torch.min}(\text{surr1}, \text{surr2})\) could result in an excessively large loss due to the minimization objective. For example: let \(\text{log\_probs} = [-0.1, -0.1, -0.1, -0.1]\), \(\text{old\_log\_probs} =  [-10, -0.2, -0.2, -5]\), and \(\text{advantages} = [-1.0, -1.0, 0.5, -0.5]\), the resulting loss values might be: \(
\text{loss} = [19930.4, 1.1, -0.5, 67.1]
\). Such large losses can make the optimization process highly unstable. Currently, many training methods remove the KL divergence constraint~\cite{meng2025mm,peng2025lmmr1}, allowing each mini-batch to perform multiple parameter updates, thereby improving data usage efficiency~\cite{hu2025reinforce,schulman2017proximal}. The former accelerates model updates, while the latter further increases the discrepancy between \(\log \pi_\theta(a_t | s_t)\) and \(\log \pi_{\theta_{\text{old}}}(a_t | s_t)\). Consequently, in these cases, the ratio between these two values can diverge significantly, leading to instability. 

\begin{algorithm}[t]
\caption{Pseudocode of PPO Loss Function in a PyTorch-like style.}
\label{alg:ppo_loss}
\definecolor{codeblue}{rgb}{0.25,0.5,0.5}
\definecolor{brown}{rgb}{0.6,0.3,0.1}  
\lstset{
  backgroundcolor=\color{white},
  basicstyle=\fontsize{7.2pt}{7.2pt}\ttfamily\selectfont,
  columns=fullflexible,
  breaklines=true,
  captionpos=b,
  commentstyle=\fontsize{7.2pt}{7.2pt}\color{codeblue},
  keywordstyle=\fontsize{7.2pt}{7.2pt},
}
\begin{lstlisting}[language=python]
# log_probs: log probabilities of the current policy
# old_log_probs: log probabilities of the previous policy
# advantages: advantage estimates for the actions
# epsilon: clipping parameter for PPO objective

# Our Pre-Clip strategy: clip the log difference to prevent large values
log_diff = log_probs - old_log_probs
log_diff = torch.clamp(log_diff, max=np.log(1e3), min=np.log(1e-3)) # similar to 10
ratio = torch.exp(log_diff)

# PPO strategy
0. ratio = (log_probs - old_log_probs).exp()  # compute the probability ratio


1. surr1 = ratio * advantages  # first surrogate objective
2. surr2 = ratio.clamp(1 - epsilon, 1 + epsilon) * advantages  # second surrogate with clipping

# The final loss is the minimum of the two surrogates
3. loss = -torch.min(surr1, surr2)  # negative loss for minimization
\end{lstlisting}
\end{algorithm}

\textbf{Instability Caused by Advantage Normalization.} In addition to the training loss, the data labels for the reward model are relatively simple, consisting of only two labels: 1 and 2, which makes them easy to learn. As a result, during training, there is a high probability that the majority of the batch will correctly predict the rewards. In extreme cases, such as a batch containing 255 rewards of 1 and 1 reward of 0, this highly imbalanced distribution, when subjected to z-Normalization, can lead to significant numerical disparities. Particularly, the advantage corresponding to the 0 reward in this example would be normalized to -15.96. A large advantage value like this can cause instability.


\section{\abbr}

\subsection{Our Training Algorithm: StableReinforce}

To overcome the drawbacks and enhance the stability of reinforcement learning training, we propose two strategies: pre-CLIP and advantage filter, which respectively remove unstable gradients and advantages that deviate excessively from the overall distribution. In terms of reward design, we introduce the consistency reward to ensure consistency between reasoning and the final answer.

\textbf{Pre-CLIP.} As shown in Algorithm~\ref{alg:ppo_loss} under the ``Our Pre-Clip strategy'', our core approach is to clip large ratios before computing the exponential of the log probability. The value of \(1e3\) is a hyperparameter that we find works well and the method is relatively insensitive to hyperparameter variations. The main purpose of this step is to mitigate the impact of noisy data on the overall training process
with log-probability clamping:
\[
\overline{\frac{\pi_\theta(a_t | s_t)}{\pi_{\theta_{\text{old}}}(a_t | s_t)}} \leftarrow \exp\left(\text{clip}\left(\log\frac{\pi_\theta}{\pi_{\theta_{\text{old}}}}, \log\delta_{\text{min}}, \log\delta_{\text{max}}\right)\right)
\]
where \(\delta_{\text{min}}=10^{-3}\), \(\delta_{\text{max}}=10^{3}\) control allowable probability ratio bounds. By clipping the ratio before applying the exponential function, we can prevent overflow issues due to excessively large differences in the ratios. Additionally, this clipping ensures that large log-probability differences are mitigated, particularly when the advantage is negative, thus maintaining training stability.

\textbf{Advantage Filter.} To prevent the influence of outliers due to the extreme imbalance in the advantage distribution, we apply the 3-sigma rule. For the standardized advantage, \( A_{\text{standardized}} = \frac{A - \mu_A}{\sigma_A} \), we retain only those advantages that fall within the range of \( [-3, 3] \)\footnote{After applying Z-normalization in the original text, the distribution becomes a standard normal distribution, meaning it has a mean of 0 and a standard deviation of 1.}. This range corresponds to values within 3 standard deviations from the mean in the original distribution, as the standardization process converts the data to z-scores (unitless measures in terms of standard deviations). In the extreme case from the previous subsection, this ensures that all samples with original rewards of 1 are selected, while extreme negative advantages are excluded.
\[
\hat{A} = \begin{cases} 
A_{\text{standardized}} & \text{if } |A_{\text{standardized}}| \leq 3 \\
0 & \text{otherwise}
\end{cases}, \quad
A_{\text{standardized}} = \frac{A - \mu_A}{\sigma_A + \epsilon}
\]


The final StableReinforce objective function with clipping applied:

\[
L^{\text{StableReinforce}}(\theta) = \frac{1}{|t|}\sum_t \left[ \min \left( \overline{\frac{\pi_\theta(a_t | s_t)}{\pi_{\theta_{\text{old}}}(a_t | s_t)}} \hat{A_t}, \, \text{clip} \left( \overline{\frac{\pi_\theta(a_t | s_t)}{\pi_{\theta_{\text{old}}}(a_t | s_t)}}, 1 - \epsilon, 1 + \epsilon \right) \hat{A_t} \right) \right],
\]
where the reward calculation and advantage estimation strategies are the same to Reinforce++.

\subsection{Remark}
In the field of RL for LLMs, recent concurrent advancements have emerged, some of which share similarities with our approach or report analogous observations. Although these methods have not been directly applied to multimodal domains or reward modeling, we provide a concise discussion in this section for comparative purposes. Notably, DAPO~\cite{yu2025dapo}, TOPR~\cite{roux2025tapered}, and Minimax-01~\cite{li2025minimax} focus on improving CLIP operations, particularly in the design of the epsilon parameter. In contrast, our approach fundamentally differs by clipping the logits ratio prior to the exponential operation. This strategy enhances numerical stability and mitigates the adverse effects of negative advantages. Similarly, Dr. GRPO~\cite{liu2025understanding} identifies the detrimental impact of advantage normalization and adopts a strategy of setting variance to 1. However, in scenarios with high original variance, this approach allows extreme values to dominate. Instead, we employ a 3-sigma filter, which preserves the benefits of z-normalization while effectively removing outliers. 


\begin{table}[t]
\centering
\caption{\textbf{Prompt template for reward model training.}}\label{tab:reward}
\renewcommand{\arraystretch}{1.5} 
\begin{tabular}{>{\raggedright\arraybackslash}p{14cm}}
\toprule
You are a highly skilled and impartial evaluator tasked with comparing two responses generated by a Large Multimodal Model for a given question. \\

- Start with a thorough, side-by-side comparative analysis enclosed within \texttt{<think>} and \texttt{</think>} tags. A tie is not permitted; you must choose a better option. \\
 - Conclude with a single numeric choice enclosed within \texttt{<answer>} and \texttt{</answer>} tags: \\
  - Output ``1'' if Response 1 is better. \\
  - Output ``2'' if Response 2 is better. \\[1em]
\textbf{Input} \\
\texttt{[Question]: \{question\}} \\
\texttt{[Response 1]: \{answer1\}} \\
\texttt{[Response 2]: \{answer2\}} \\[1em]
\textbf{Output Format (strictly follow)} \\
\texttt{<think>Your detailed comparative analysis</think><answer>1/2</answer>} \\ 
\bottomrule
\end{tabular}
\end{table}

\subsection{Reward Function and Training Data}

Inspired by DeepSeek-R1~\cite{deepseekai2025deepseekr1incentivizingreasoningcapability}, we aim to directly use RL to guide the reward model in generating the best analysis content, in order to produce high-quality model output comparisons. As a result, the prompt format in Table \ref{tab:reward} transforms the reward modeling task into a straightforward rule-based reinforcement learning problem. By defining the model's output format, we only need to define our reward functions to complete the training process:

\begin{itemize}
    \item \textbf{Formatting Reward.} The model’s output must adhere to a specific format of `<think> </think><answer> </answer>`, which encourages the model to reason before generating the final output. This ensures that the model reflects on the reasoning process before making its final decision, enhancing both the quality and interpretability of the generated content.
    \item \textbf{Result Reward.} The model’s generated final result must align with human preferences. This primarily involves ensuring that the model's output ranking labels are consistent with those of human experts, enhancing the overall usefulness and credibility.
\end{itemize}

\textbf{Inconsistency Between Reasoning and Results.} However, simply following existing work~\cite{deepseekai2025deepseekr1incentivizingreasoningcapability,yu2025dapo} in our setting has led to unexpected results. During training, we observe discrepancies between the model's reasoning and its final answer. For example, the reasoning might conclude that response 2 is better but the model outputs answer 1, as seen in \textit{<think>... response 2 is better</think><answer>1</answer>}. This inconsistency arises because, we provide no supervision for the reasoning process and only score based on the outcome. When a sample demonstrates poor reasoning but produces the correct answer, this pattern is inadvertently reinforced, leading the model to believe that reasoning and the final answer are not necessarily linked. Consequently, the model may learn to generate correct answers without a coherent reasoning process. This could even result in the model treating the reasoning process as irrelevant, or worse, outputting repetitive content or random noise. To address this issue, we introduce an additional component, Qwen2.5-VL-7B-Instruct, as a supervisor to verify whether the reasoning and the final result are consistent. This addition helps ensure that the reasoning process and output align well, introducing the following reward function:

\begin{itemize}
 \item \textbf{Consistency Reward.} The model’s final result must be consistent with its intermediate reasoning process. This function ensures that the final answer is directly derived from the model’s reasoning process, rather than being generated in isolation from the reasoning steps.
\end{itemize}
    
Integrating the consistency reward as a separate reward and combining it with the previous two reward functions can lead to a situation where the model, despite selecting the wrong answer, may still receive a high overall reward due to the consistency component. This could result in the model overly prioritizing consistency. To mitigate this issue, the final reward is designed as follows:
\[
\text{Final Reward} = \text{Result Reward} \times (1 + 0.5 \times \text{Consistency Reward}) + 0.5 \times \text{Formatting Reward}.
\]
This ensures that the consistency reward is only taken into account when the result is correct, thereby preventing the model from excessively favoring consistency in cases where the outcome is incorrect.

\textbf{Dataset Construction.} As shown in Table \ref{tab:data}, we sample preference data from multiple existing datasets for training. To ensure data quality and diversity, we sample all instances from the human-annotated dataset MM-RLHF, and an additional 100,000 samples from other multimodal preference datasets. The final dataset is termed R1-Reward-200k, which combines these diverse instances to create a robust training foundation for our model. We then randomly shuffle the data to ensure a balanced ratio of answers 1 and 2 (1:1), preventing the model from favoring a specific answer. Each sample consists of a quadruple: (question, answer 1, answer 2, ground truth choice).

\textbf{Long-Cot Cold Start.} Since MLLMs are not initially trained for reward modeling tasks, directly using MLLMs for reinforcement learning training yields poor and unstable results. Therefore, we first use GPT-4o as an annotator for each sample in the R1-Reward-200k dataset, utilizing the prompts from Table \ref{tab:reward} to construct SFT data. The temperature is set to 0, with a maximum of 3 attempts. We also record how many attempts GPT-4o needs to generate a final response that is judged to be correct (i.e., the same to the ground truth choice). This SFT training phase teaches the model the basic format and familiarizes it with the reward modeling task.

\textbf{RL Training Data.} We use all samples from the SFT phase with at least 2 attempts, as well as samples where GPT-4o fails to produce the correct answer after three attempts. These samples exhibit smaller differences between answer 1 and answer 2, making them more difficult.

\begin{figure}
    \centering
    \includegraphics[width=\linewidth]{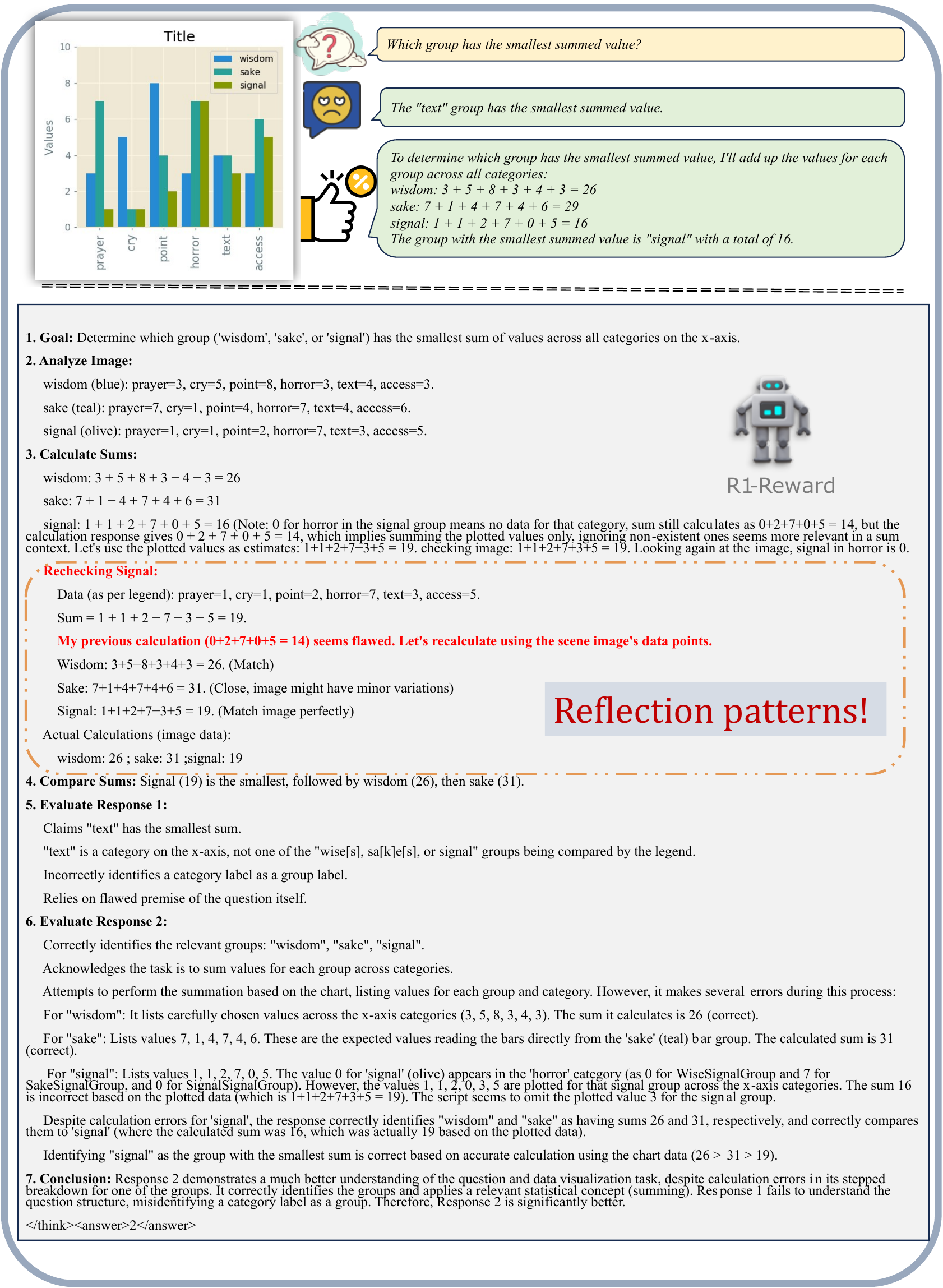}
    \vspace{-0.4cm}
    \caption{\textbf{An example of the R1-Reward output.} R1-Reward shows ``human-like'' self-reflective
thought process, which is also called ``Aha moment''~\cite{deepseekai2025deepseekr1incentivizingreasoningcapability}.}
    \vspace{-0.3cm}
    \label{fig:output_example}
     
\end{figure}

\begin{table}[]
\caption{\textbf{Summary of datasets used for training}, including the category (text or image), dataset name, the number of original samples, and the number of samples selected for final training.}
\label{tab:data}
\resizebox{\textwidth}{!}{%
\begin{tabular}{cccccc}
\rowcolor[HTML]{EFEFEF} 
\toprule
\textbf{Dataset} & \textbf{\# Original} & \textbf{\# Sample} & \textbf{Dataset} & \textbf{\# Original} & \textbf{\# Sample} \\
RLAIF-V~\cite{yu2024rlaifvopensourceaifeedback} & 74,802 & \multirow{4}{*}{100k} & MM-RLHF-Long~\cite{zhang2025mm} & 41,163 & 41,163 \\
VL-Feedback~\cite{2023vlfeedback} & 80,258 &  & MM-RLHF-Short~\cite{zhang2025mm} & 46,281 & 46,281 \\
POVID~\cite{zhou2024aligningmodalitiesvisionlarge} & 17,184 &  & MM-RLHF-Mcq~\cite{zhang2025mm} & 8,306 & 8,306 \\
WildVision-Battle~\cite{lu2024wildvision} & 10,383 &  & MM-RLHF-Safety~\cite{zhang2025mm} & 9,990 & 9,990 \\ \bottomrule
\end{tabular}%
}
 \vspace{-0.2cm}
\end{table}


\vspace{-0.1cm}
\section{Experiments}
\vspace{-0.1cm}

\textbf{Implementation Details.}  Both SFT and RL experiments are conducted on 4$\times$H800 (80G) GPUs. The SFT phase trains for 1 epoch and takes approximately 8 hours, while the RL phase trains for 5 epochs and takes 12 hours. We use QwenVL-2.5-7B-Instruct as the base model for training. During the SFT phase, the learning rate is set to 1e-5, and the batch size is set to 256. We use the OpenRLHF~\cite{hu2024openrlhf} framework for RL. The training batch size is set to 128, and the rollout batch size is set to 256. The learning rate is set to 1e-6, and the initial KL coefficient is set to 0.

\textbf{Baseline Algorithm.} At the algorithmic level, we primarily compare two entities: the reward model and MM-RLHF-Reward \cite{zhang2025mm}. For the former, we replace the language head of the base LLM with a two-layer MLP that outputs a float value as the reward. Training is done using a binary classification loss. For the latter, in addition to the traditional binary classification loss, an additional critic loss is required. Specifically, the model first outputs an evaluation of the candidate, and then, based on the evaluation, the reward head provides the reward value.

\textbf{Baseline Models.} For multimodal reward models, we consider GPT-4o-mini (2024-07-18), Claude-3.5-Sonnet (2024-06-22), Gemini-1.5-Flash (2024-09-24), GPT-4o (2024-08-06), Gemini-1.5-Pro (2024-09-24), Gemini-2.0-Flash-Exp, SliME~\cite{zhang2024benchmarking}, VITA-1.5~\cite{fu2025vita}, LLaVA-OneVision-7B-ov~\cite{li2024llava}, Qwen2-VL-7B~\cite{wang2024qwen2}, Molmo-7B~\cite{deitke2024molmo}, InternVL2/3-8B~\cite{chen2023internvl,zhu2025internvl3}, LLaVA-Critic-8B~\cite{xiong2024llava}, Llama-3.2-11B~\cite{INF-ORM-Llama3.1-70B}, Pixtral-12B~\cite{agrawal2024pixtral}, Molmo-72B~\cite{deitke2024molmo}, Qwen2-VL-72B~\cite{wang2024qwen2}, NVLM-D-72B~\cite{dai2024nvlm}, MM-RLHF-Reward-7B~\cite{zhang2025mm}, Llama-3.2-90B~\cite{INF-ORM-Llama3.1-70B} and IXC-2.5-Reward~\cite{zang2025internlm} as comparison points. 

\textbf{Evaluation Benchmarks and Metrics.} The multimodal benchmark consists of the VL-Reward Bench~\cite{li2024vlrewardbenchchallengingbenchmarkvisionlanguage}, Multimodal RewardBench~\cite{yasunaga2025multimodal} and the MM-RLHF-Reward Bench~\cite{zhang2025mm}. VL-Reward Bench includes two evaluation metrics: Overall Accuracy and Macro Average Accuracy. \textit{Overall Accuracy} measures the percentage of model decisions that align with human preferences, while \textit{Macro Average Accuracy} calculates the mean accuracy across various task categories, addressing task distribution imbalance. Multimodal RewardBench serves as a comprehensive benchmark for evaluating reward models. It covers six key areas: general correctness, preference, knowledge, reasoning, safety, and visual question answering (VQA). This benchmark includes 5,000 annotated triplets, each consisting of a (multimodal prompt, chosen response, rejected response) pair. The MM-RLHF-Reward Bench also features two evaluation metrics: 1. \textit{Traditional Accuracy (Acc)}: This metric assesses the proportion of cases where the model correctly identifies the preferred response. 2. \textit{Acc+}: This measures the proportion of cases where the model correctly ranks all response pairs for a given sample. This metric emphasizes the model's ability to handle challenging cases, such as those with small ranking differences or hard-to-distinguish pairs.

\definecolor{front-color}{HTML}{FDEFF5}
\begin{table}[]
\caption{\textbf{VLReward Bench.} Performance comparison of our reward model (R1-Reward) with existing open-source and private counterparts.}
\label{tab:vl_rewardbench}
\resizebox{\textwidth}{!}{%
\begin{tabular}{@{}lcccccc@{}}
\toprule
\textbf{Models} & \textbf{\#Param} & \textbf{General} & \textbf{Hallucination} & \textbf{Reasoning} & \textbf{Overall Acc} & \textbf{Macro Acc} \\ \midrule \rowcolor[HTML]{FDEFF5} 
\multicolumn{7}{c}{\texttt{Proprietary Models}} \\
GPT-4o-mini (2024-07-18) & - & 41.70 & 34.50 & 58.20 & 41.50 & 44.80 \\
Claude-3.5-Sonnet (2024-06-22) & - & 43.40 & 55.00 & 62.30 & 55.30 & 53.60 \\
Gemini-1.5-Flash (2024-09-24) & - & 47.80 & 59.60 & 58.40 & 57.60 & 55.30 \\
GPT-4o (2024-08-06) & - & 49.10 & 67.60 & \underline{70.50} & 65.80 & 62.40 \\
Gemini-1.5-Pro (2024-09-24) & - & 50.80 & {72.50} & 64.20 & {67.20} & 62.50 \\
Claude-3.7-Sonnet & - & {68.08} & 70.70 & 60.81 & 66.31 & 66.53 \\
\hline \rowcolor[HTML]{FDEFF5} 
\multicolumn{7}{c}{\texttt{\texttt{Open-Source Models}}} \\
VITA-1.5 & 7B & 18.55 & 8.93 & 22.11 & 16.48 & 16.53 \\
SliME & 7B & 7.23 & 27.09 & 18.60 & 19.04 & 17.64 \\
LLaVA-OneVision-7B-ov & 7B & 32.20 & 20.10 & 57.10 & 29.60 & 36.50 \\
Molmo-7B & 7B & 31.10 & 31.80 & 56.20 & 37.50 & 39.70 \\
InternVL2-8B & 8B & 35.60 & 41.10 & 59.00 & 44.50 & 45.20 \\
LLaVA-Critic-8B & 8B & 54.60 & 38.30 & 59.10 & 41.20 & 44.00 \\
Llama-3.2-11B & 11B & 33.30 & 38.40 & 56.60 & 42.90 & 42.80 \\
Pixtral-12B & 12B & 35.60 & 25.90 & 59.90 & 35.80 & 40.40 \\
Molmo-72B & 72B & 33.90 & 42.30 & 54.90 & 44.10 & 43.70 \\
Qwen2-VL-72B & 72B & 38.10 & 32.80 & 58.00 & 39.50 & 43.00 \\
NVLM-D-72B & 72B & 38.90 & 31.60 & 62.00 & 40.10 & 44.10 \\ \hline
\rowcolor[HTML]{FDEFF5} \multicolumn{7}{c}{\texttt{\texttt{Reward Models}}} \\
MM-RLHF-Reward & 7B & 45.04 & 50.45 & 57.55 & 50.15 & 51.01 \\
Llama-3.2-90B & 90B & 42.60 & 57.30 & 61.70 & 56.20 & 53.90 \\
IXC-2.5-Reward & 7B & \textbf{84.70} & 62.50 & 62.90 & 65.80 & {70.00 }\\ \hline\rowcolor[HTML]{FDEFF5} 
\multicolumn{7}{c}{Ours} \\
R1-Reward & 7B & {63.84} & \underline{85.71} & {64.78} & \underline{71.92 }& \underline{71.44} \\
\textit{Voting@15} & 7B & \underline{66.32} &\textbf{ 89.06} & \textbf{73.70} & \textbf{76.46} & \textbf{76.36}
\\ \bottomrule
\end{tabular}%
}

\end{table}
\subsection{Main Results}

We evaluate the performance of R1-Reward on three common multimodal reward model benchmarks. On the VLReward Bench (Table~\ref{tab:vl_rewardbench}), R1-Reward achieves the best overall performance, with an average accuracy of $71.92\%$. This represents a roughly $9.3$\% improvement in overall accuracy compared to the previous best open-source model, IXC-2.5-Reward. Notably, IXC-2.5-Reward trains on more than 1 million samples, while our training data consists of 200k samples, highlighting a significant improvement in data efficiency. In comparison to other open-source models, R1-Reward demonstrates a larger margin of improvement. Among closed-source models, Gemini-1.5-Pro performs the best, but R1-Reward outperforms it across all dimensions, further demonstrating its superiority.

On the Multimodal Reward Bench (Table~\ref{tab:tab_mm_reward}), R1-Reward achieves the best performance across all dimensions, with a $14.3\%$ improvement over the previous state-of-the-art. It is worth noting that the Multimodal Reward Bench is derived from over ten existing benchmarks and reconstructed into a unified set, with minimal overlap with our training data. This further demonstrates R1-Reward’s remarkable generalization ability across different datasets.

The MM-RLHF-Reward Bench (Table~\ref{tab:tab_mmrlhf_reward})  presents a higher level of difficulty, particularly when directly utilizing language models as reward models. The best-performing model, Claude-3.7-Sonnet, achieves an accuracy of $65\%$ on the Acc+ metric. Existing reward models perform well, with IXC-2.5-Reward surpassing an Acc+ score of $50\%$, while the top reward model, MM-RLHF-Reward, exceeds $60\%$. However, MM-RLHF-Reward is trained on a dataset that closely aligns with the distribution of this benchmark, which limits its generalization ability. As a result, its performance on the VL Reward Benchmark is suboptimal. In contrast, R1-Reward demonstrates balanced performance across all benchmarks. Moreover, when performing voting on five sampled results, its accuracy reaches $85.3\%$, and when sampling 15 times, it reaches $86.47\%$—significantly outperforming existing models.

\begin{figure*}[t]
\subfigure[MM-RLHF Reward Bench]{
\begin{minipage}[t]{0.32\linewidth}
\centering
 \includegraphics[width=\linewidth]{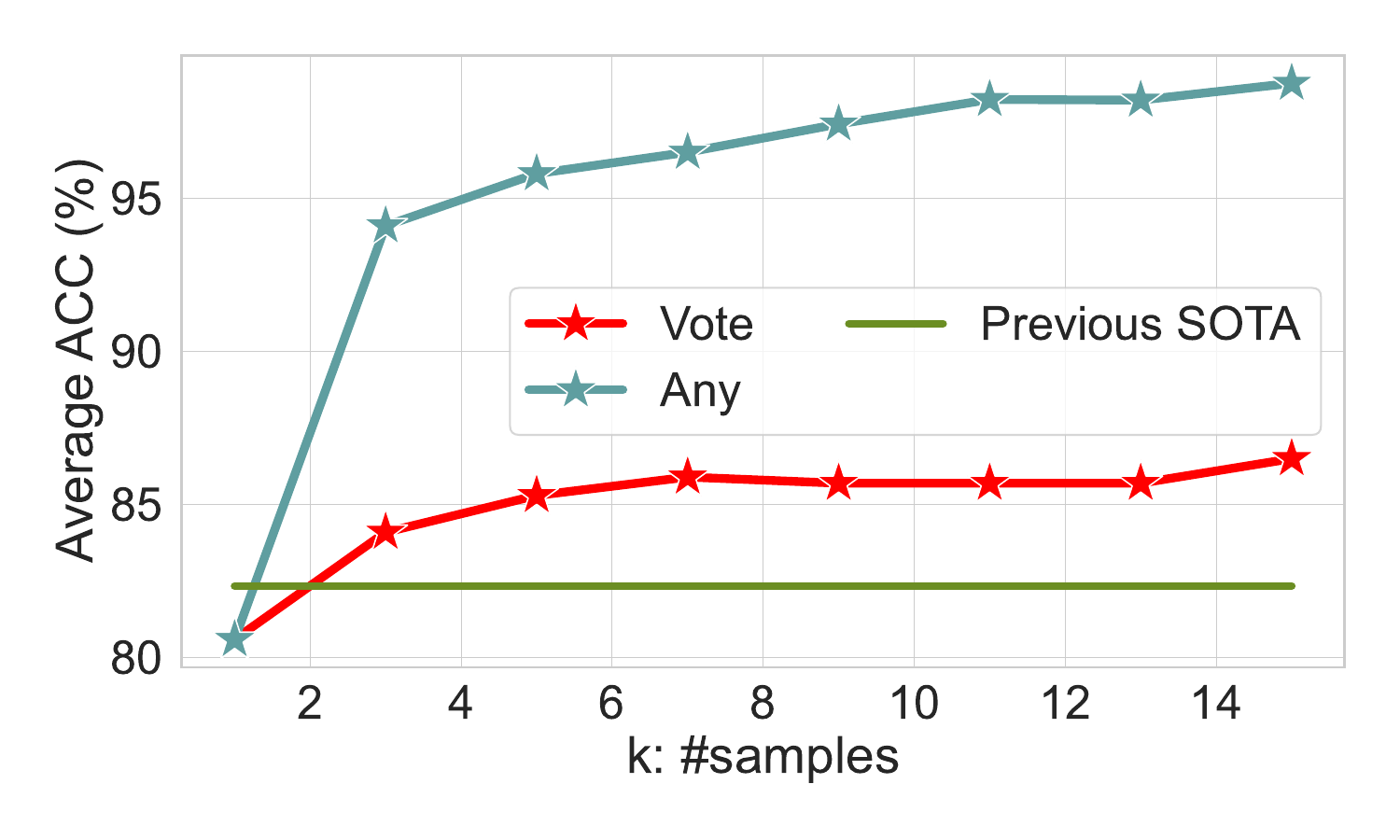}
\end{minipage}%
}%
\subfigure[VL Reward Bench]{
\begin{minipage}[t]{0.32\linewidth}
 \includegraphics[width=\linewidth]{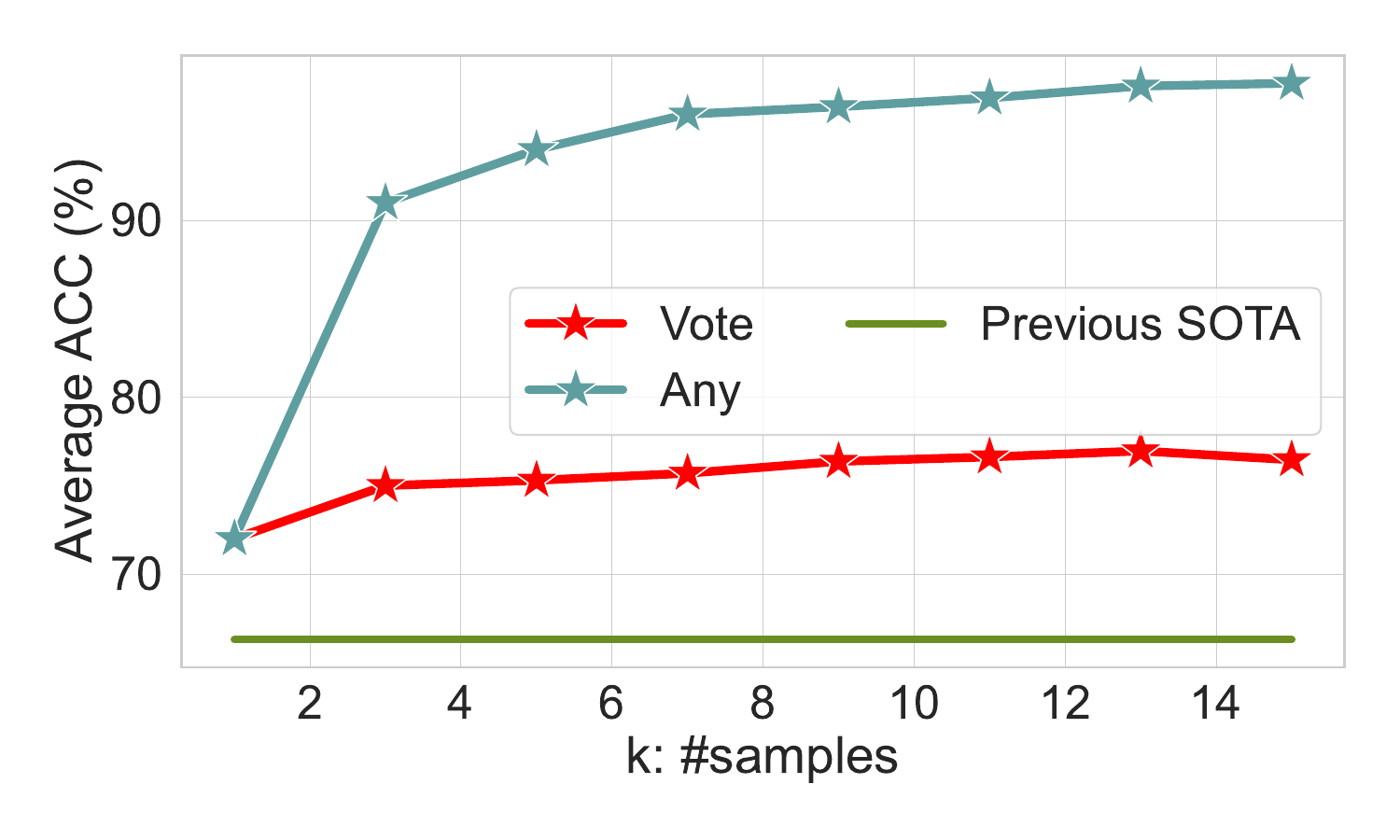}
\end{minipage}%
}
\subfigure[Multimodal Reward Bench]{
\begin{minipage}[t]{0.32\linewidth}
\centering
 \includegraphics[width=\linewidth]{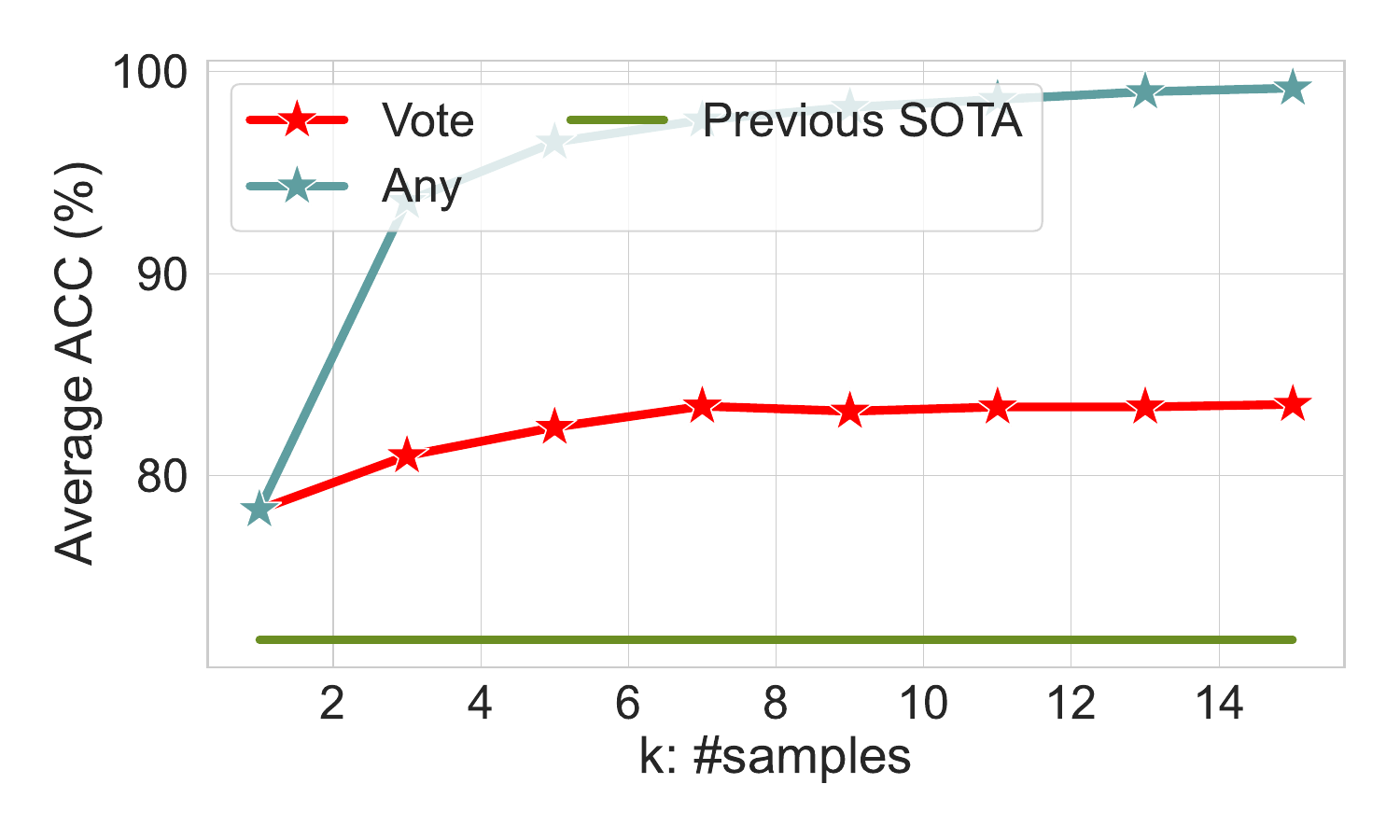}
\end{minipage}%
}%
\centering
\vspace{-0.2cm}
\caption{Inference-time performance scaling of R1-Reward on three benchmarks: (a) MM-RLHF Reward Bench, (b) VL Reward Bench, and (c) Multimodal Reward Bench. Accuracy is measured using two aggregation strategies as the number of inference samples (K) increases: ``Majority Vote'' and ``Any Correct''. The ``Any Correct'' strategy (successful if at least one of the K samples is correct) is highly sensitive to K, while the ``Majority Vote'' strategy shows a more gradual improvement. Performance is compared against the previous SOTA result for each benchmark.}
\label{fig:test_tme}
\end{figure*}

\textbf{Test-Time Scaling.} In Figure \ref{fig:test_tme}, we explore whether increasing the number of samples can lead to improved performance. It is important to note that the temperature is set to $1.0$, which causes the result with a single sample ($k=1$) to slightly differ from the main results (which use greedy decoding by default). As the number of samples increases, the model's performance improves consistently. ``Vote'' refers to a majority-voting strategy, while ``Any'' counts as correct if at least one of the sampled results is correct. At $k=15$, the accuracy of ``Any'' approaches $100\%$, indicating that the R1-Reward has the potential to perfectly classify all samples; however, additional data or training strategies are needed to fully unlock this potential. Moreover, the ``Vote'' results demonstrate a significant advantage over previous state-of-the-art models, with a more noticeable improvement when $k < 5$. The benefits from increasing the number of samples gradually diminish as more samples are added. This highlights the potential of R1-Reward in test-time scaling.

\subsection{Ablations and Analysis}

\textbf{R1-Reward Demonstrates High Data Efficiency.} In Table~\ref{tab:ablation}, we compare the performance of a traditional reward model (using a two-layer MLP as the reward head) and MM-RLHF-Reward (which first generates a critic and then generates the reward) trained on the same dataset. For MM-RLHF-Reward, the training data must include an evaluation for each response. To achieve this, we use GPT-4o to generate corresponding evaluations for each sample, which may be slightly less accurate than the human annotations used in the original work. All the models' backbones are Qwen2.5-VL-7B-Instruct. As shown in the table, the traditional reward model, when trained with only 200K data samples, performs poorly. In most cases, MM-RLHF outperforms the traditional reward model. Its superior performance in the ``hallucination'' dimension is likely due to the generated critic. Comparing these two baselines, the reinforcement learning-based approach significantly enhances the reward modeling capabilities, even with the same amount of data. Moreover, our SFT approach shows advantages over both the traditional reward model and MM-RLHF-Reward. We believe this is primarily due to that we allow direct comparison of two responses during the scoring process, whereas existing methods score responses independently before comparing them.

\begin{figure*}[t]
\subfigure[]{
\begin{minipage}[t]{0.49\linewidth}
\centering
 \includegraphics[width=\linewidth]{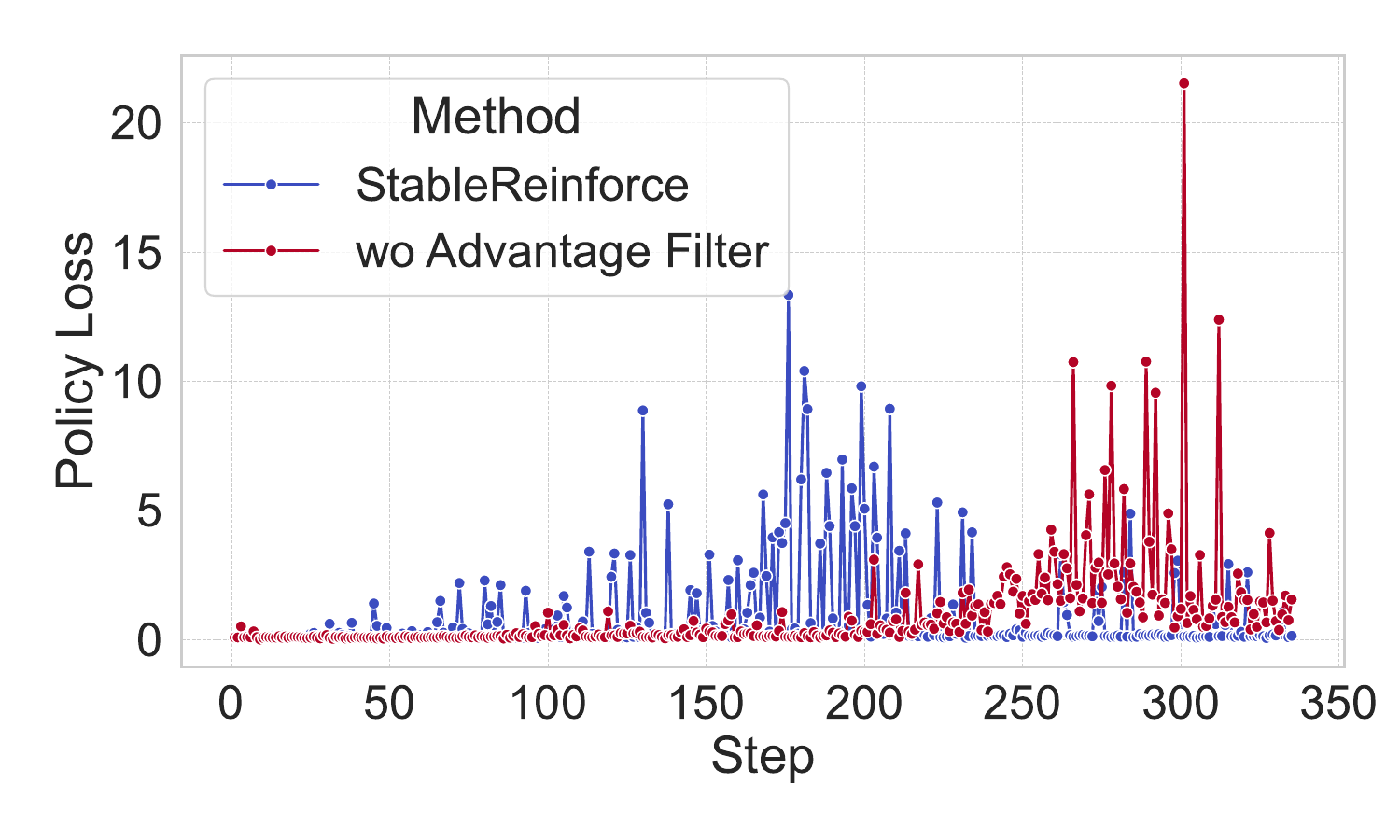}
\end{minipage}%
}%
\subfigure[]{
\begin{minipage}[t]{0.49\linewidth}
 \includegraphics[width=\linewidth]{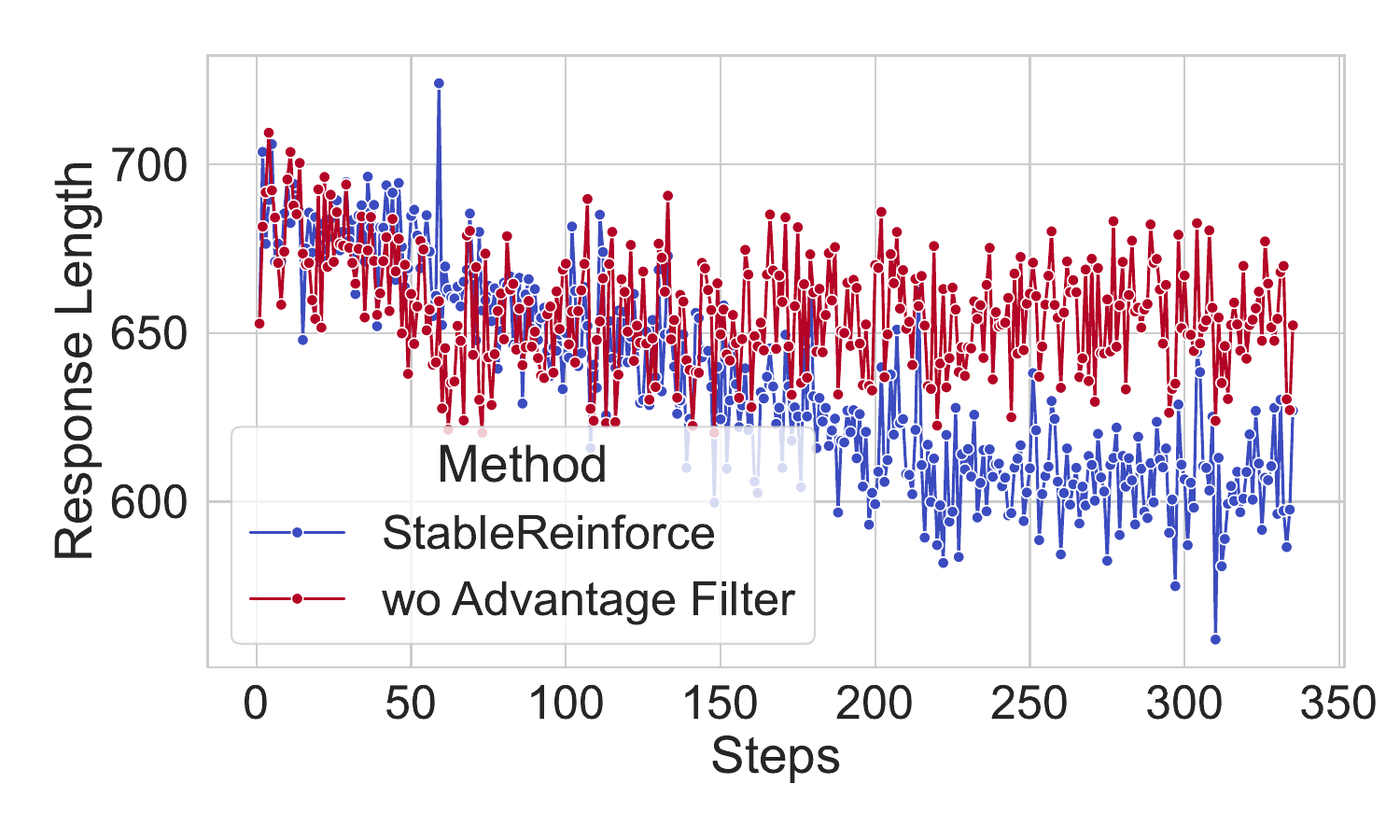}
\end{minipage}%
}
\subfigure[]{
\begin{minipage}[t]{0.49\linewidth}
\centering
 \includegraphics[width=\linewidth]{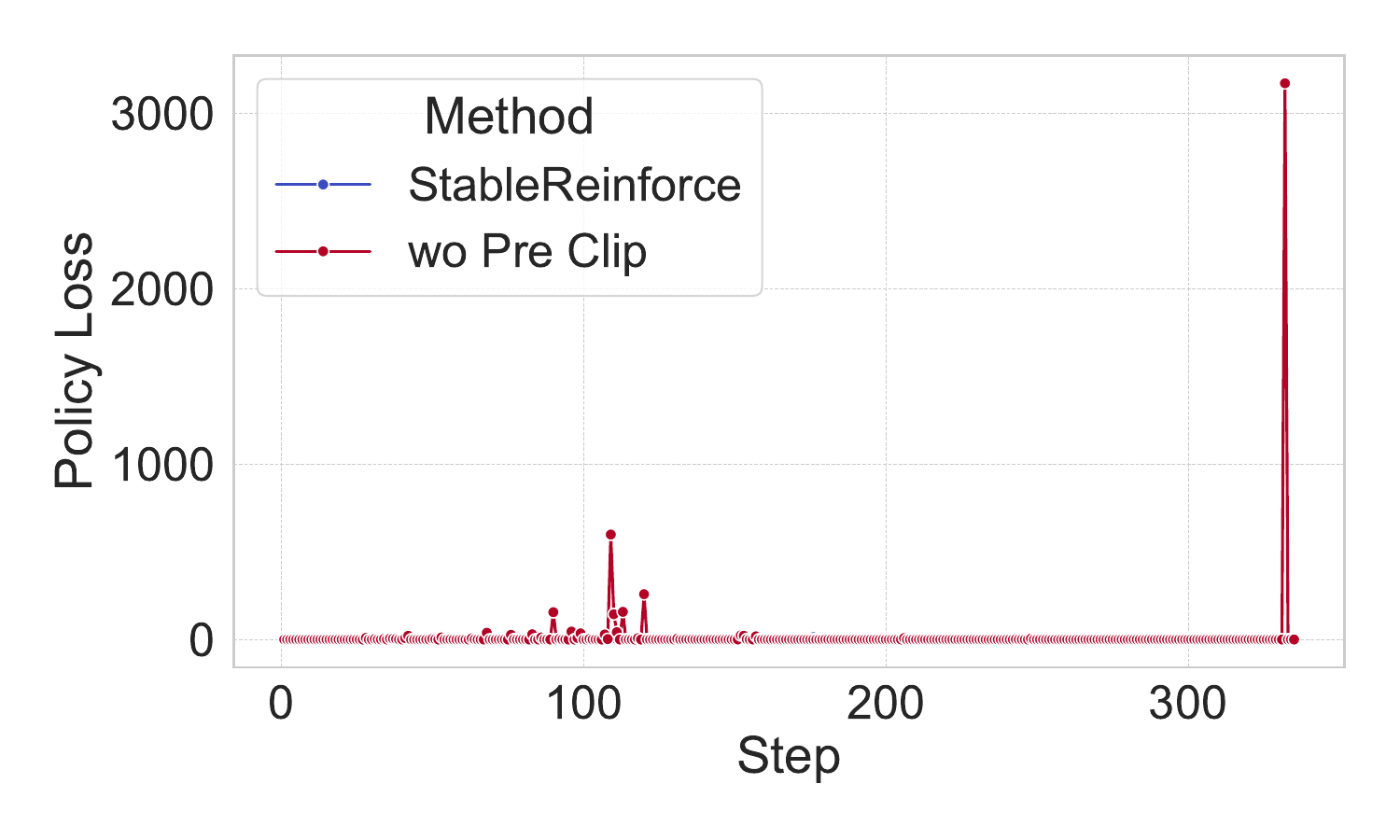}
\end{minipage}%
}%
\subfigure[]{
\begin{minipage}[t]{0.49\linewidth}
 \includegraphics[width=\linewidth]{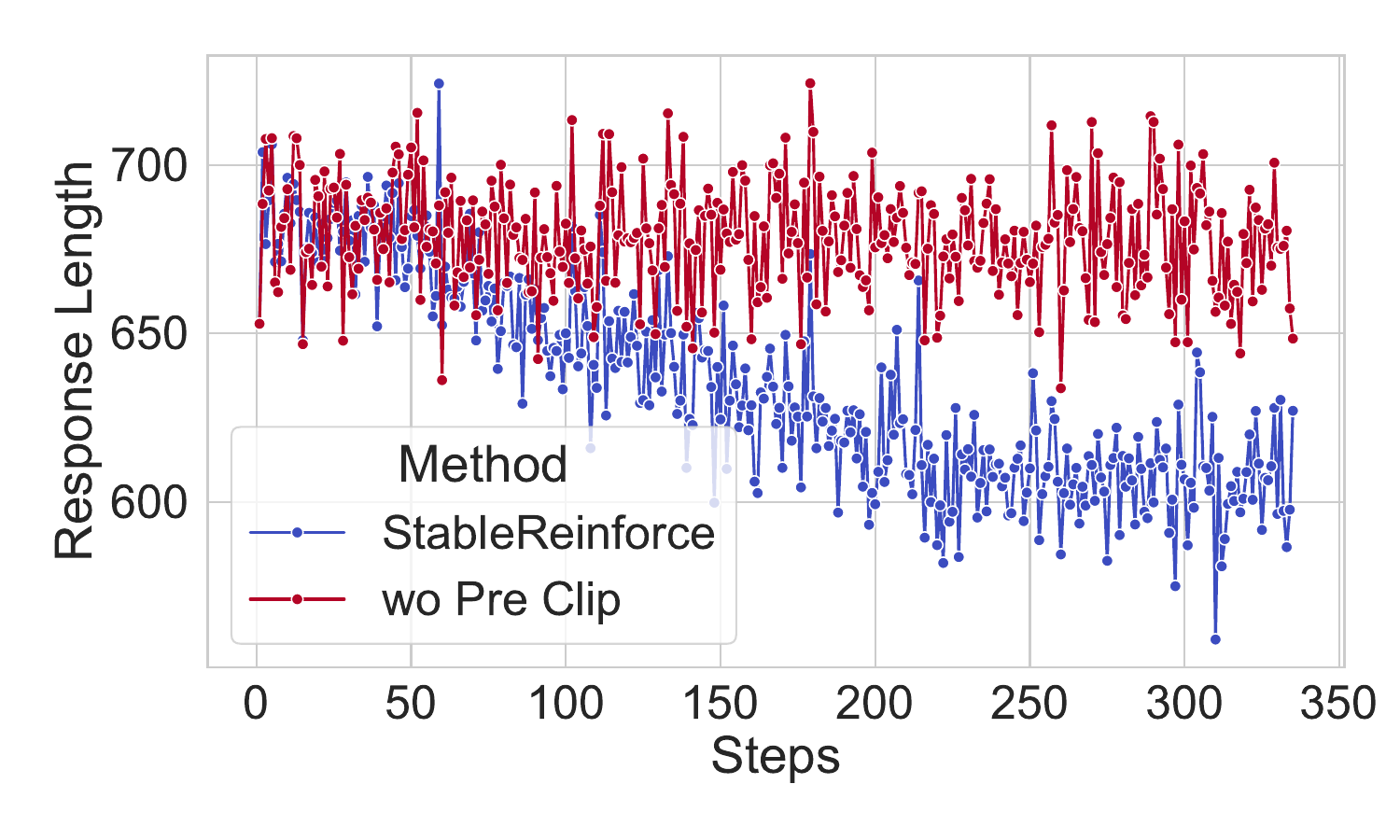}
\end{minipage}%
}
\centering
\vspace{-0.2cm}
\caption{\textbf{Ablation studies of the StableReinforce algorithm}, evaluating the impact of different components on policy loss and model response length. The subfigures compare the performance of the algorithm with and without specific components: (a) and (b) show results when the advantage filter is removed; (c) and (d) when Pre Clip is removed. Each method is evaluated in terms of policy loss and response length over training steps.
}
\label{fig:ablation_adv}
\end{figure*}

\textbf{Ablation Studies of the StableReinforce Algorithm.} We examine the impact of each component of the StableReinforce algorithm on the training process and final results. First, we emphasize the necessity of the Consistency Reward Function. Removing this function results in significant hallucination behaviors across different algorithms, making it challenging to achieve stable evaluation outcomes. Additionally, directly applying the Reinforce++ algorithm causes the model to crash, with the loss becoming NaN and the response length reaching the preset maximum length, while the output consists entirely of garbled text. In Table~\ref{tab:ablation} and Figure~\ref{fig:ablation_adv}, we present the effects of removing each module on final performance and changes in training dynamics. We observe that the Advantage Filter and Pre-Clip modules primarily ensure training stability by effectively removing outliers from the loss. Removing any of these components results in decreased final accuracy, reduced training stability, and the model's output length failing to converge to shorter values.

\begin{figure}
    \centering
    \includegraphics[width=\linewidth]{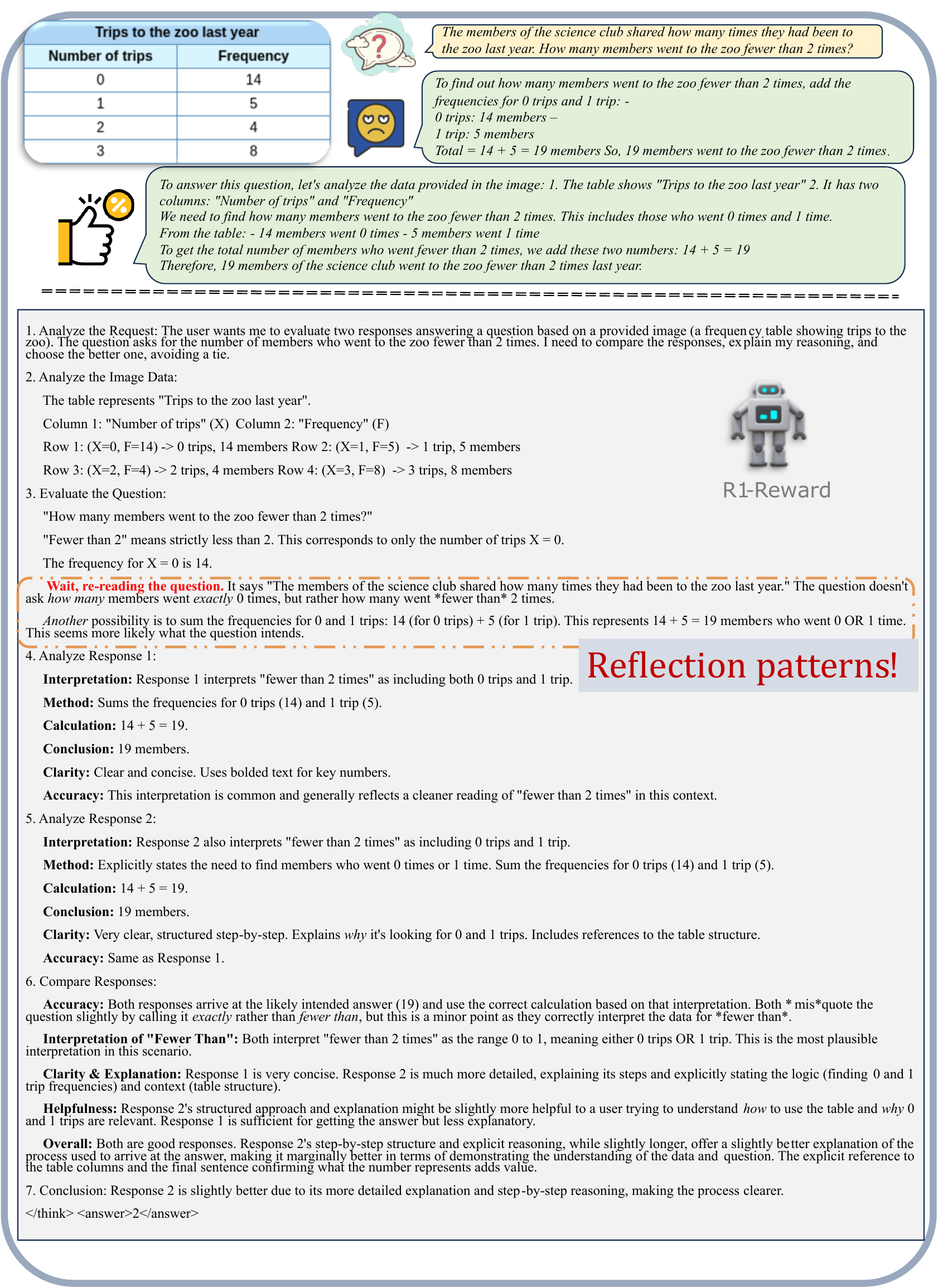}
    \vspace{-0.4cm}
    \caption{\textbf{An example of the R1-Reward output.}}
    \vspace{-0.3cm}
    \label{fig:output_example2}
     
\end{figure}

\textbf{Aha Moment of R1-Reward.} Through our task design and reward function formulation, the R1-Reward model effectively learns the reward modeling task structure during the SFT phase. Following reinforcement learning, it reduces the length of reasoning to enhance efficiency. Visual examples of the model’s output appear in Figures~\ref{fig:output_example} and~\ref{fig:output_example2}. The model autonomously learns a process to assess response quality. It first defines the goal, analyzes the image, attempts to solve the problem, and provides an answer. Based on this, the model evaluates Response 1 and Response 2, compares the two outputs, and gives a final ranking. Simultaneously, the model demonstrates different reflection patterns. In Figure~\ref{fig:output_example}, the model encounters an error in its calculation, but after rechecking the bar chart, it recognizes the mistake and recalculates to obtain the correct result. In Figure~\ref{fig:output_example2}, the model misunderstands the problem. However, after outputting ``Wait, re-reading the question,'' it re-reads the question, eventually understands it correctly, and determines the correctness of the answer.

\vspace{-0.1cm}
\section{Conclusion}
\vspace{-0.1cm}
In this paper, we introduce R1-Reward, a MRM trained using the StableReinforce algorithm. We demonstrate that RL can be effectively applied to reward modeling, significantly enhancing its performance. Our approach addresses key challenges, including training instability, the advantage normalization limitation, and inconsistencies between reasoning and results. By incorporating techniques such as pre-clipping, advantage filtering, consistency reward and a a progressive difficulty training strategy, StableReinforce stabilizes training and improves model performance. Experiments show that R1-Reward outperforms SOTA models on several multimodal reward model benchmarks, with significant improvements in accuracy and data efficiency.

Furthermore, R1-Reward demonstrates excellent test-time scaling capabilities, and paves the way for future research on integrating reinforcement learning into MRMs. Looking ahead, there are still many areas to explore in RL for reward modeling. For example, we only test a simple majority voting strategy for test-time scaling; more advanced methods could potentially improve performance further~\cite{liu2025inference}. Additionally, improving training strategies to further enhance the foundational capabilities of reward models is also a meaningful open problem.

\begin{table}[]
\caption{\textbf{Multimodal Reward Bench. }Performance comparison of our reward model (R1-Reward) with existing open-source and proprietary counterparts.}
\label{tab:tab_mm_reward}
\resizebox{\textwidth}{!}{%
\begin{tabular}{lccccccccc}
\toprule
\multirow{2}{*}{\textbf{Model}} & \multirow{2}{*}{\#\textbf{Param}} & \multirow{2}{*}{\textbf{Overall}} & \multicolumn{2}{c}{\textbf{General}} & \multirow{2}{*}{\textbf{Knowledge}} & \multicolumn{2}{c}{\textbf{Reasoning}} & \multirow{2}{*}{\textbf{Safety}} & \multirow{2}{*}{\textbf{VQA}} \\ \cmidrule{4-5}\cmidrule{7-8}
 &  &  & \textbf{Correctness} & \textbf{Preference} &  & \textbf{Math} & \textbf{Coding} &  &  \\ \hline \rowcolor[HTML]{D9D2E9}
\multicolumn{10}{c}{\texttt{Proprietary Models}} \\
GPT-4o & - & 70.8 & 62.6 & {69.0} & 72.0 & 67.6 & 62.1 & 74.8 & \textbf{87.2} \\
Gemini 1.5 Pro & - & {71.9} & {63.5} & 67.7 & 66.3 & {68.9} & 55.5 & {94.5} & \textbf{87.2} \\
Claude 3.5 Sonnet & - & 71.5 & 62.6 & 67.8 & {73.9} & 68.6 & {65.1} & 76.8 & 85.6 \\  
Claude 3.7 Sonnet &  & {71.9} & 58.4 & 60.7 & \textbf{78.1} & {76.3} & {71.3} & 72.0 & {86.8}  \\
\rowcolor[HTML]{D9D2E9}\hline
\multicolumn{10}{c}{\texttt{Open-Source Models}} \\
SliME & 8B & 42.0 & 42.3 & 52.2 & 47.5 & 43.5 & 35.3 & 19.1 & 53.8 \\
VITA-1.5 & 7B & 53.6 & 55.6 & 54.3 & 52.5 & 51.9 & 52.8 & 58.1 & 50.0 \\
Llama-3.2-Vision-Instruct & 11B & 51.2 & 57.8 & 65.8 & 55.5 & 50.6 & 51.7 & 20.9 & 55.8 \\
Molmo-D-0924 & 7B & 52.9 & 56.8 & 59.4 & 54.6 & 50.7 & 53.4 & 34.8 & 60.3 \\
Llama-3.2-Vision-Instruct & 90B & 61.2 & 60.0 & 68.4 & 61.2 & 56.3 & 53.1 & 52.0 & 77.1 \\
InternVL-3 & 8B & 63.6 & 59.6 & 61.6 & 60.5 & 65.1 & 56.6 & 59.3 & 82.3 \\
Qwen-2-VL & 72B & 70.9 & 56.4 & 62.3 & 70.2 & 73.3 & 58.9 & 90.1 & 85.3\\  \rowcolor[HTML]{D9D2E9} \hline
\multicolumn{10}{c}{\texttt{Reward Models}} \\
MM-RLHF-Reward & 7B & 67.1 & 61.7 & 67.5 & 54.3 & 58.4 & 57.9 & 92.9 & 76.8 \\ 
IXC-2.5-Reward & 7B & 66.6 & 60.7 & 64.2 & 56.8 & 63.0 & 50.5 & 89.9 & 81.1 \\
 \rowcolor[HTML]{D9D2E9} \hline
\multicolumn{10}{c}{\texttt{Ours}} \\
R1-Reward & 7B & \underline{82.2} & \underline{77.5} & \underline{74.0} & \underline{74.9} & \textbf{83.1} & \underline{79.6} & \textbf{99.6} & {86.5} \\ 
Voting@15 & 7B & \textbf{83.3} & \textbf{78.0} & \textbf{77.2} & 74.6 & \underline{81.3} & \textbf{85.8} & \underline{99.4} & \underline{87.0} 
\\
\bottomrule
\end{tabular}%
}
\end{table}

\begin{table}[]
\caption{\textbf{MM-RLHF-Reward Bench. }Performance comparison of our reward model (R1-Reward) with existing open-source and proprietary counterparts.}
\label{tab:tab_mmrlhf_reward}
\resizebox{\textwidth}{!}{%
\begin{tabular}{@{}lcccccccc@{}}
\toprule
\textbf{Models} & \textbf{\#Param} & \textbf{Mcq} & \textbf{Long} & \textbf{Short} & \textbf{Safety} & \textbf{Video} & \textbf{Acc} & \textbf{Acc+} \\ \midrule\rowcolor[HTML]{F5FFFA}
\multicolumn{9}{c}{\texttt{Proprietary Models}} \\ 
Gemini-2.0-Flash-Exp & - & 33.33 & 45.94 & 67.64 & 43.75 & 32.00 & 44.71 & 13.04 \\
GPT-4o (2024-08-06) & - & {64.28} & 78.37 & 44.11 & 56.25 & 40.00 & 58.23 & 26.01 \\
Claude-3.5-Sonnet (2024-06-22) & - &  {64.28} & 67.56 & 55.88 & 65.62 & 60.00 & 62.94 & 26.11 \\
Claude-3.7-Sonnet & - & 66.67 & {91.89} & \textbf{91.18} & \textbf{87.50} & 76.00 & \underline{82.35} & \underline{65.22}\\
\hline\rowcolor[HTML]{F5FFFA}
\multicolumn{9}{c}{\texttt{Open-Source Models}} \\ 
SliME & 8B & 23.81 & 10.81 & 14.71 & 12.50 & 7.52 & 17.10 & 1.76 \\
VITA-1.5 & 7B & 24.97 & 21.62 & 11.76 & 18.75 & 12.40 & 20.58 & 2.78 \\
Intern-VL-3 & 8B & 35.71 & 56.76 & 23.53 & 37.50 & 32.00 & 37.65 & 6.52 \\
NVLM-D-72B & 72B & 42.85 & 32.43 & 8.82 & 50.00 & 40.00 & 34.70 & 6.52 \\
Llama-3.2-90B & 90B & 19.04 & 35.13 & 38.23 & 50.00 & 40.00 & 35.29 & 10.86 \\
Qwen2-VL-72B & 72B & 45.23 & 62.16 & 47.05 & 46.88 & 36.00 & 48.23 & 13.04 \\
\hline\rowcolor[HTML]{F5FFFA}\multicolumn{9}{c}{\texttt{\texttt{Reward Models}}} \\
IXC-2.5-Reward & 7B & 52.38 &  {91.89} & 67.65 & 62.50 & \textbf{88.00} & 71.18 & {50.00} \\
MM-RLHF-Reward & 7B & \underline{83.00} & \underline{97.00} & {74.00} & {69.00} & \textbf{88.00} & {82.00} & {63.00} \\\hline\rowcolor[HTML]{F5FFFA}
\multicolumn{9}{c}{\texttt{Ours}}  \\
{R1-Reward} & 7B & {80.95} & {89.19} & \underline{82.35} & {75.00} & {72.00} & {80.59} & {54.35} \\ 
Voting@15 & 7B & \textbf{83.33} & \textbf{97.30} & \textbf{91.18} & \underline{78.12} & \underline{80.00} & \textbf{86.47 }& \textbf{67.39 }\\
\bottomrule
\end{tabular}%
}

\end{table}

\begin{table}[htbp] 
\caption{\textbf{Evaluation results} on VL Reward Bench comparing different models and training setups, including baselines, models trained with R1-Reward-200K, and ablation studies (\texttt{Ours}).}
\label{tab:ablation} 
\centering 
\resizebox{\textwidth}{!}{%
\begin{tabular}{llcccc} 
\toprule
\multicolumn{1}{c}{\multirow{2}{*}{\textbf{\# Data}}} & \multicolumn{1}{c}{\multirow{2}{*}{\textbf{Models}}} & \multicolumn{4}{c}{\textbf{VL-Reward Bench}} \\ \cmidrule(lr){3-6} 
\multicolumn{1}{c}{} & \multicolumn{1}{c}{} & \textbf{General} & \textbf{Hallucination} & \textbf{Reasoning} & \textbf{Overall Acc} \\ 
\hline\Gray 
\multicolumn{6}{c}{\texttt{Baselines}} \\ 
More than 1M & IXC-2.5-Reward & 84.70 & 62.50 & 62.90 & 65.80 \\ 
MM-RLHF-120K & MM-RLHF-Reward & 45.04 & 50.45 & 57.55 & 50.15 \\
\hline\Gray 
\multicolumn{6}{c}{\texttt{Trained by R1-Reward-200K}} \\ 
R1-Reward-200K & Reward Model & 56.71 & 56.03 & 48.67 & 56.41 \\
R1-Reward-200K & MM-RLHF-Reward & 61.01 & 62.28 & 59.30 & 60.80 \\
\Gray\hline 
\multicolumn{6}{c}{\texttt{Ours}} \\ 
R1-Reward-200K & StableReinforce & 63.84 & 85.71 & 64.78 & 71.92 \\
R1-Reward-200K & wo advantage & 63.43 & 77.45 & 62.38 & 68.96 \\
R1-Reward-200K & wo pre-clip & 62.06 & 77.44 & 61.23 & 67.36 \\
R1-Reward-200K & Reinforce++ & \multicolumn{4}{c}{Collapse} \\
R1-Reward-200K & Only Long-Cot SFT & 59.92 & 72.27 & 60.01 & 64.80 \\
\bottomrule
\end{tabular}%
}

\end{table}

\bibliographystyle{unsrt}
\bibliography{neurips_2024}

\clearpage
\newpage
\appendix

\end{document}